\documentclass[10pt,twocolumn,letterpaper]{article}

\usepackage[pagenumbers]{cvpr} % To force page numbers, e.g. for an arXiv version

\usepackage{graphicx}
\usepackage{amsmath}
\usepackage{amssymb}
\usepackage{booktabs}

\usepackage{multirow}
\usepackage{tabularx}
\usepackage{caption}
\usepackage{subcaption}
\usepackage{cite}
\usepackage{url}
\usepackage[hidelinks]{hyperref}

\usepackage[capitalize]{cleveref}
\crefname{section}{Sec.}{Secs.}
\Crefname{section}{Section}{Sections}
\Crefname{table}{Table}{Tables}
\crefname{table}{Tab.}{Tabs.}

\def\cvprPaperID{*****} % *** Enter the CVPR Paper ID here
\def\confName{CVPR}
\def\confYear{2023}

\begin{document}

%%%%%%%%% TITLE - PLEASE UPDATE
\title{MultiEarth 2023 -- Multimodal Learning for Earth and Environment \\ Workshop and Challenge}

\author{Miriam Cha\textsuperscript{1}, Gregory Angelides\textsuperscript{1}, Mark Hamilton\textsuperscript{2}, Andy Soszynski\textsuperscript{1}, Brandon Swenson\textsuperscript{3},\\ Nathaniel Maidel\textsuperscript{3}, Phillip Isola\textsuperscript{2}, Taylor Perron\textsuperscript{2}, Bill Freeman\textsuperscript{2} \\ \\
\textsuperscript{1}MIT Lincoln Laboratory, [miriam.cha, gregangelides, andrew.soszynski]@ll.mit.edu\\
\textsuperscript{2}MIT, [markth, phillipi, perron, billf]@mit.edu \\
\textsuperscript{3}DAF MIT AI Accelerator, [brandon.swenson.2, nathaniel.maidel]]@us.af.mil}

\maketitle

%%%%%%%%% ABSTRACT
\begin{abstract}
The Multimodal Learning for Earth and Environment Workshop (MultiEarth 2023) is the second annual CVPR workshop aimed at the monitoring and analysis of the health of Earth ecosystems by leveraging the vast amount of remote sensing data that is continuously being collected. The primary objective of this workshop is to bring together the Earth and environmental science communities as well as the multimodal representation learning communities to explore new ways of harnessing technological advancements in support of environmental monitoring. The MultiEarth Workshop also seeks to provide a common benchmark for processing multimodal remote sensing information by organizing public challenges focused on monitoring the Amazon rainforest. These challenges include estimating deforestation, detecting forest fires, translating synthetic aperture radar (SAR) images to the visible domain, and projecting environmental trends. This paper presents the challenge guidelines, datasets, and evaluation metrics. Our challenge website is available at  \url{https://sites.google.com/view/rainforest-challenge/multiearth-2023}.
\end{abstract}

\section{Introduction}

The Amazon rainforest is one of the most biodiverse regions on Earth, home to an incredible array of plant and animal species. It covers over 5.5 million square kilometers, accounting for more than half of the total rainforest remaining on Earth \cite{juan_2016}. Due to this fact, the rainforest plays a crucial role in regulating the Earth's climate, absorbing and storing large amounts of carbon dioxide from the atmosphere, earning it the nickname ``Lungs of the Planet'' \cite{bbc_amazon}. However, the rainforest is at risk due to deforestation, fires, climate change, and other environmental effects. World Wildlife Fund (WWF) estimates that if the current rate of deforestation continues, around 40\% of the Amazon rainforest will be lost by 2050 \cite{wwf}. Should the current trends persists, the Amazon might not exist in a few generations, leading to catastrophic repercussions for all life on the planet. Therefore, it is more crucial than ever to comprehend how the Amazon rainforest's ecosystem is evolving over time. 

The Multimodal Learning for Earth and Environment Challenge (MultiEarth 2023) is the second annual CVPR workshop aimed at analyzing the environmental changes in the Amazon rainforest by leveraging multimodal remote sensing data. Remote sensing offers a unique vantage point, providing wide coverage and periodic observations that are otherwise challenging to achieve on the ground. Combining the capabilities of active sensors, such as synthetic aperture radar (SAR), with passive optical sensors enhances forest monitoring efforts. SAR's independence from weather and lighting conditions enables image acquisitions in the presence of clouds and darkness, overcoming limitations of passive sensors (\eg optical). A key component of the MultiEarth 2023 challenge involves the integration of SAR and optical images to conduct comprehensive assessment of the forest dynamics, at any time and under any weather conditions.

In recent years, numerous studies have explored the application of remote sensing data for assessing and monitoring forest ecosystems \cite{ijgi9100580,lima_2019,Coelho_2021,Ngadze_2020}. Passive, optical sensors such as the U.S. Geological Survey’s Landsat \cite{landsat}, NASA’s Moderate Resolution Imaging Spectroradiometer (MODIS) \cite{modis}, the European Space Agency’s Sentinel-2 \cite{sentinel}, and China–Brazil Earth Resources Satellite (CBERS) \cite{CBERS}, have been employed to capture multispectral data, enabling the detection of forest landcover changes, deforestation, and forest fragmentation. 

% While considerable research has been devoted, the analyses typically rely on passive, optical sensors, which require an unobstructed and illuminated view of the scene.  

Several researchers have curated benchmark datasets for multimodal remote senseing data, encompassing images from passive and active sensors. Sumbul \etal~\cite{Sumbul_2021, Sumbul_2019} developed the BigEarthNet dataset, which contains 590k pairs of Sentinel-1 and Sentinel-2 images covering ten European countries. Another dataset, SpaceNet6 \cite{shermeyer2020spacenet}, offers 4k quad-pol X-band SAR and optical image pairs with a resolution of 900$\times$900 at 0.5m over Rotterdam, Netherlands. Additionally, MDAS \cite{mdas_2023} compiled a dataset for the city of Germany, comprising SAR and optical images alongside a digital surface model (DSM). However, a benchmark dataset focusing on the Amazon rainforest is still lacking. To facilitate future research, we released the MultiEarth 2022 dataset \cite{cha2022multiearth}, which comprises a continuous time series of Sentinel-1, Sentinel-2, Landsat 5 and Landsat 8 images, accompanied by corresponding deforestation labels. As an enhancement this year, we have included additional downstream labels indicating areas affected by forest fire.

MultiEarth 2023 introduces four sub-challenges focusing on the interpretation and analysis of the rainforest at any time and any weather conditions, as follows: 1) fire detection, 2) deforestation estimation, 3) environmental trend prediction, and 4) SAR-to-EO image translation sub-challenges. In the followings, we describe the challenge problems and the guidelines for participating.

\begin{itemize}
\item \textbf{Fire Detection Sub-Challenge}: The fire detection sub-challenge is a new task focusing on automatic analysis of wildfire patterns using a combination of multiple sensor outputs. Participants will be given the multimodal remote sensing dataset along with the information on areas affected by fire. Performance is measured on the following metrics: pixel accuracy, F1 score, and Intersection over Union (IoU). A detailed description of this sub-challenge can be found in Section~\ref{subsec:fire}. 

\item \textbf{Deforestation Estimation Sub-Challenge}: Fire detection and deforestation estimation are closely related as fires are a common cause and effect of deforestation. This sub-challenge aims to identify areas where deforestation has occurred. We provide multimodal remote sensing dataset along with the deforestation label maps. The same metrics used in the fire detection sub--challenge will be used to evaluate the submissions on the deforestation estimation sub-challenge. More information can be found in Section~\ref{subsec:dec}.

\item \textbf{Environmental Trend Prediction Sub-Challenge}: 
The ability to predict the appearance of the Amazon rainforest using historical remote sensing images is a powerful tool that can help us gain insights into how the Amazon is changing and better understand the factors driving these changes. In this sub-challenge, participants are required to predict the appearance of an overhead image of the Amazon at a specified \{location, modality\} query with a multi-year projection given historical multimodal remote sensing imagery. Performance is measured on the following visual metrics: Peak Signal-to-Noise Ratio (PSNR), Structural Similarity Index Measure (SSIM) \cite{ssim}, Learned Perceptual Image Patch Similarity (LPIPS) \cite{lpips}, and Fr\'{e}chet Inception Distance (FID) \cite{fid}. A detailed challenge description is provided in Section~\ref{subsec:mcc}. 

\item \textbf{SAR-to-EO Image Translation Sub-Challenge}: 
Translating SAR images to the optical domain enables easily interpretable wide-area imaging that is invariant to weather and lighting conditions. For this sub-challenge, participants will predict a set of possible cloud-free corresponding optical images given an input SAR image.  A multimodal aligned imagery dataset is provided (e.g. $[\mathbf{x} ,[\mathbf{y}_1,\mathbf{y}_2,...,\mathbf{y}_N]]$) where an input SAR image $\mathbf{x}$ is paired to a set of ground truth optical images $[\mathbf{y}_1,\mathbf{y}_2,...,\mathbf{y}_N]$. This is to reflect that a single SAR image can be correctly mapped to multiple optical images. Performance is evaluated based on $\sum_j \min_i \Arrowvert f(\mathbf{x})_i - \mathbf{y}_j \Arrowvert$ where $f(\cdot)$ is a prediction of $\mathbf{y}$ given $\mathbf{x}$, and $f$ may make multiple predictions indexed by $i$. A detailed guideline of the sub-challenge is provided in Section~\ref{subsec:i2ic}. 
\end{itemize}

% \begin{figure}[t]
% \centering 
% \includegraphics[width=.35\textwidth]{Images/tapajos.png} 
% \caption{Study area in the Amazon, bounded by (4.39$^{\circ}$ S, 55.2$^{\circ}$ W), (4.39$^{\circ}$ S, 54.48$^{\circ}$ W), (3.33$^{\circ}$ S, 54.48$^{\circ}$ W) and (3.33$^{\circ}$ S, 55.2$^{\circ}$ W).}
% \label{fig:tapajos} 
% \end{figure}  

% Please add the following required packages to your document preamble:
% \usepackage{multirow}
\begin{table*}[t]
\small
\centering 
\begin{tabular}{ccccccc}
\hline
\multirow{2}{*}{\textbf{Data}} & \multirow{2}{*}{\textbf{Time}} & \multirow{2}{*}{\textbf{Bands}}                                                                                        & \multirow{2}{*}{\textbf{Resolution (m)}} & \multirow{2}{*}{\textbf{\# Images}} & \multicolumn{2}{c}{\textbf{Link}}                                                                                             \\ \cline{6-7} 
                               &                                &                                                                                                                        &                                          &                                     & \multicolumn{1}{l}{NetCDF}                                    & \multicolumn{1}{l}{ZIP}                                       \\ \hline
Sentinel-1                     & 2014-2021                      & VV, VH                                                                                                                 & 10                                       & 780,706                             & \href{https://rainforestchallenge.blob.core.windows.net/multiearth2023-dataset-final/sent1_train.nc} {\color{blue}{link}}                                                             & \href{https://rainforestchallenge.blob.core.windows.net/multiearth2023-dataset-final/sent1_train.zip} {\color{blue}{link}}                                                               \\ \hline
Sentinel-2                     & 2018-2021                      & \begin{tabular}[c]{@{}c@{}}B1, B2, B3, B4, B5,\\ B6, B7, B8, B8A, B9,\\ B11, B12, QA60\end{tabular}                    & 10                                       & 5,241,119                           & \begin{tabular}[c]{@{}c@{}}\href{https://rainforestchallenge.blob.core.windows.net/multiearth2023-dataset-final/sent2_b1-b4_train.nc} {\color{blue}{link1}}\\ \href{https://rainforestchallenge.blob.core.windows.net/multiearth2023-dataset-final/sent2_b5-b8_train.nc} {\color{blue}{link2}}\\ \href{https://rainforestchallenge.blob.core.windows.net/multiearth2023-dataset-final/sent2_b9-b12_train.nc} {\color{blue}{link3}}\end{tabular} & \begin{tabular}[c]{@{}c@{}}\href{https://rainforestchallenge.blob.core.windows.net/multiearth2023-dataset-final/sent2_b1-b4_train.zip} {\color{blue}{link1}} \\ \href{https://rainforestchallenge.blob.core.windows.net/multiearth2023-dataset-final/sent2_b5-b8_train.zip} {\color{blue}{link2}} \\ \href{https://rainforestchallenge.blob.core.windows.net/multiearth2023-dataset-final/sent2_b9-b12_train.zip} {\color{blue}{link3}} \end{tabular} \\ \hline
Landsat 5                      & 1984-2012                      & \begin{tabular}[c]{@{}c@{}}SR\_B1, SR\_B2, SR\_B3,\\ SR\_B4, SR\_B5, ST\_B6,\\ SR\_B7, QA\_PIXEL\end{tabular}          & 30                                       & 3,432,640                           & \href{https://rainforestchallenge.blob.core.windows.net/multiearth2023-dataset-final/landsat5_train.nc} {\color{blue}{link}}                                                          & \href{https://rainforestchallenge.blob.core.windows.net/multiearth2023-dataset-final/landsat5_train.zip} {\color{blue}{link}}                                                           \\ \hline
Landsat8                       & 2013-2021                      & \begin{tabular}[c]{@{}c@{}}SR\_B1, SR\_B2, SR\_B3,\\ SR\_B4, SR\_B5, SR\_B6,\\ SR\_B7, ST\_B10, QA\_PIXEL\end{tabular} & 30                                       & 2,099,025                           & \href{https://rainforestchallenge.blob.core.windows.net/multiearth2023-dataset-final/landsat8_train.nc} {\color{blue}{link}}                                                            & \href{https://rainforestchallenge.blob.core.windows.net/multiearth2023-dataset-final/landsat8_train.zip} {\color{blue}{link}}                                                             \\ \hline
\end{tabular}
\caption{Overview of the MultiEarth dataset}
\label{tab:data}
\end{table*}

\section{Datasets}
\subsection{Multimodal Remote Sensing Dataset}
\label{subsec:multimodal_data}

While the increasing volume of remote sensing data presents exciting opportunities for insights and discoveries, it also creates significant challenges in processing and analyzing the data at a large scale. This year, we provide documentation and an API along with a tutorial that can aid in loading and filtering the MultiEarth data.  To facilitate future research in multimodal remote sensing, we have released our code at \url{https://github.com/MIT-AI-Accelerator/multiearth-challenge}.

One update we've made in this year's challenge is providing the MultiEarth dataset in a NetCDF format \cite{netcdf}. NetCDF is a format that is specifically designed for storing  and managing scientific data. It has several advantages over a general-purpose compression format like ZIP when it comes to storing and processing large amounts of image data, allowing for efficient access to specific subsets of the data, which can be important when dealing with datasets that are too large to load into memory all at once. Additionally NetCDF includes support for metadata that can be used to describe the data in a standardized way, making it easier to share and understand. Our API  includes methods to load datasets specific to each challenge from the NetCDF files. Additionally, we also provide some simple utilities to aid in retrieving spatially and temporally aligned TIFF images extracted from ZIP if that format is desired.

Similar to MultiEarth 2022, participants will receive a multimodal remote sensing dataset that consists of Sentinel-1, Sentinel-2, Landsat 5, and Landsat 8 as reported in Table~\ref{tab:data}. Sentinel-1 uses a synthetic aperture radar (SAR) instrument, which collects in two polarization bands: VV (vertical transmit/vertical receive) and VH (vertical transmit/horizontal receive). Sentinel-2, Landsat 5, and Landsat 8 use optical instruments, which measure in spectral bands in the visible and infrared spectra. We also include in the dataset the associated layers with cloud quality for Sentinel-2, Landsat 5, and Landsat 8 (i.e. QA60 and QA\_PIXEL). Detailed band designations for each sensor can be found in the Google Earth Engine Data Catalog\footnote{\scriptsize Sentinel-1:  \url{https://developers.google.com/earth-engine/datasets/catalog/COPERNICUS_S1_GRD} \\
Sentinel-2: \url{https://developers.google.com/earth-engine/datasets/catalog/COPERNICUS_S2_SR}\\
Landsat 5: \url{https://developers.google.com/earth-engine/datasets/catalog/LANDSAT_LT05_C02_T1_L2} \\
Landsat 8: \url{https://developers.google.com/earth-engine/datasets/catalog/LANDSAT_LC08_C02_T1_L2}
}. This year's challenge filters out some of the flawed images from the dataset used in the previous year. 

\begin{figure*}[t]
     \centering
     \begin{subfigure}[b]{0.25\textwidth}
         \centering
         \includegraphics[width=\textwidth]{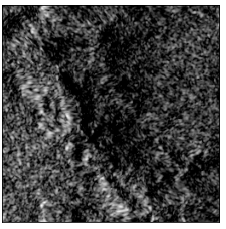}
         \caption{Sentinel-1}
         \label{fig:sentinel1}
     \end{subfigure}
     \hfill
     \begin{subfigure}[b]{0.25\textwidth}
         \centering
         \includegraphics[width=\textwidth]{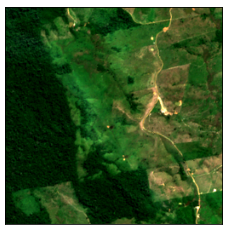}
         \caption{Sentinel-2}
         \label{fig:sentinel2}
     \end{subfigure}
     \hfill
     \begin{subfigure}[b]{0.25\textwidth}
         \centering
         \includegraphics[width=\textwidth]{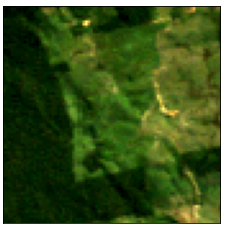}
         \caption{Landsat 5}
         \label{fig:landsat5}
     \end{subfigure}
     \hfill
     \begin{subfigure}[b]{0.25\textwidth}
         \centering
         \includegraphics[width=\textwidth]{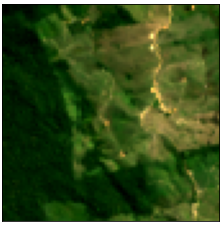}
         \caption{Landsat 8}
         \label{fig:landsat8}
     \end{subfigure}
    \hfill
     \begin{subfigure}[b]{0.25\textwidth}
         \centering
         \includegraphics[width=\textwidth]{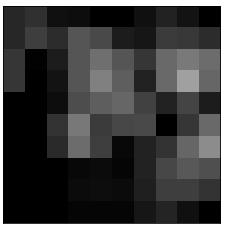}
         \caption{Fire}
         \label{fig:fire}
     \end{subfigure}
         \hfill
     \begin{subfigure}[b]{0.25\textwidth}
         \centering
         \includegraphics[width=\textwidth]{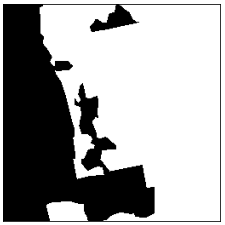}
         \caption{Deforestation}
         \label{fig:defor}
     \end{subfigure}
        \caption{Example images from the MultiEarth dataset. (a) Sentinel-1 VV, (b) Sentinel-2 RGB (B4, B3, B2), (c) Landsat 5 RGB (SR\_B3, SR\_B2, SR\_B1), (d) Landsat 8 RGB (SR\_B4, SR\_B3, SR\_B2), (e) Fire ConfidenceLevel, and (f) Deforestation label map.}
        \label{fig:example_chips}
\end{figure*}

Satellite images from this dataset are acquired using Google's Earth Engine platform. To account for the difference in pixel resolutions, a 256$\times$256 image is extracted for Sentinel-1 and Sentinel-2, whereas for Landsat 5 and Landsat 8, an 85$\times$85 image is extracted. In order to achieve geospatial alignment with the Sentinel images, the Landast images can be upsampled to a resolution of 256$\times$256. We show example images extracted from the same latitude and longitude coordinates
(LAT/LON: -4.21/-54.98) but on different dates in Figure~\ref{fig:example_chips}.

The dataset can be downloaded from the links provided in Table~\ref{tab:data}. The dataset is freely available for development, research, and educational purposes, with additional information provided in the Google Earth Engine License Agreement\footnote{\url{https://earthengine.google.com/terms/}}.

\subsection{Labels}
\label{subsec:label}
\noindent\textbf{Fire.}
%\label{subsubsec:fire_data}
We use the global burned area maps derived from MODIS by the ESA Fire Disturbance Climate Change Initiative (CCI) project \cite{fire_cci}. The Fire dataset consists of monthly collections spanning from 2001 to 2020, with a resolution of 250 meters. Each image in the dataset is of size 10$\times$10 and contains four bands: BurnDate, ConfidenceLevel, LandCover, and ObservedFlag. BurnDate indicates the estimated day of the year when the burn was first detected. ConfidenceLevel includes the probability of identifying a pixel as being burned. LandCover categorizes the land cover category
of burned pixels. ObservedFlag holds information on unprocessed pixels. More details of each bands can be found in the FireCCI51 data catalog\footnote{ FireCCI51: \url{https://developers.google.com/earth-engine/datasets/catalog/ESA_CCI_FireCCI_5_1}}. We define a pixel as burned when the ConfidenceLevel is greater than 50. Burned areas are assigned a value of 1, while unburned areas are assigned a value of 0. An example fire image (ConfidenceLevel) scaled to [0,100] is shown in Figure~\ref{fig:example_chips} (e). We released 1,503,360 training images. Note that challenge participants are not allowed to download additional training data.

%\subsubsection{Deforestation}
\noindent\textbf{Deforestation.}
The labeled deforestation dataset is manually labeled using cloud-free monthly mosaic satellite images from Planet \cite{planet} over 11 time slices from 2016 to 2021 (\ie 08/2016; 07, 08/2017; 06, 08/2018; 07, 08/2019; 06, 08/2020; and 05, 08/2021). August 2021 is labeled by the authors and the remaining ten are labeled by the team at Scale AI \cite{scaleai}. The deforested pixels are denoted as 1's and the forested or other pixels are denoted as 0's. The images have a fixed size of 256$\times$256 pixels. We release 17,215 training image chips in total. This dataset corresponds to the one utilized in MultiEarth 2022, and additional information can be found in the MultiEarth 2022 white paper \cite{cha2022multiearth}.

\section{Challenge Tasks}

\subsection{Fire Detection Sub-Challenge}
\label{subsec:fire}
\noindent\textbf{Problem Definition.} Numerous research studies have found that climate change  has already led to an escalation in the duration of wildfire seasons, the frequency of wildfires, and the amount of land destroyed by fires \cite{westerling_2006, running_2006, nobre_2016}. The fire detection sub-challenge aims to better understand the Amazon rainforest destroyed by forest fires. By analyzing data from multiple sources, including optical and synthetic aperture radar (SAR) images, participants can track changes in the frequency, intensity, and location of forest fires over time. This information can help identify areas that are particularly susceptible to forest fire, as well as areas where the risk of forest fires is increasing. Automated fire detection system will provide valuable insights into how environmental changes, such as climate change, are impacting the rainforest.

\noindent\textbf{Data.} Participants will use the multimodal remote sensing dataset described in Section~\ref{subsec:multimodal_data} to predict the binary labels indicating burned and unburned areas.

\noindent\textbf{Metrics.} Performance is measured on the following metrics: pixel accuracy, F1 score, and Intersection over Union (IoU). 

\noindent\textbf{Submission Format.} We provide two separate guidelines for NetCDF and TIFF file users. \\
\noindent \textbf{(NetCDF File Users):} We will provide a NetCDF file that includes an array of 1,000 target test samples. Each sample will contain longitude, latitude, collection date, and a data source (‘Fire’) related to a test target. Functionality for retrieving images related to these test targets is provided in our code repository. \\ 
\noindent \textbf{(TIFF File Users):} We will provide a csv file that includes 1,000 test queries as a list of lists \ie [[${\text{lon}_0, \text{lat}_0, \text{date}_0, \text{modality}_0}],\text{…}, [\text{lon}_{999}, \text{lat}_{999}, \text{date}_{999}$, $\text{modality}_{999}]]$. Each test query is in the format [lon, lat, date, modality]. For example, $[-55.15, -4.11, 2021\_08\_01,$ $\text{fire}]$ will represent fire\_-55.15\_-4.11\_2021\_08\_01.tiff. To have a consistent naming convention for the date (\ie year\_month\_day), we add a nominal day label of “\_01” to all fire estimation test queries. Functionality for retrieving TIFF images related to these queries is provided in our code repository. 

Participants will submit in total 1,000 10$\times$10 binary masks, one 10$\times$10 binary mask for each input test query. 0's in the binary mask will represent unburned areas and 1's will represent fire/burned areas. These masks will be compared to the FireCCI51 data’s ConfidenceLevel band with a threshold of 50. The output files should have dtype of uint8 and should use the naming convention Fire\_Lon\_Lat\_Date.tiff. For example, the expected output file name for test query [-55.15, -4.11, 2021\_08\_01, fire] is fire\_-55.15\_-4.11\_2021\_08\_01.tiff. Imagery related to the input test queries will be made available for the purpose of generating the requested output. 

% For the test input, 1,000 test queries will be provided as a list of lists \ie [[${\text{lon}_0, \text{lat}_0, \text{date}_0, \text{modality}_0}],\text{…}, [\text{lon}_{4999}, \text{lat}_{4999}, \text{date}_{4999}$, $\text{modality}_{4999}]]$. Each test query is in the format [lon, lat, date, modality]. For example, $[-55.15, -4.11, 2021\_08\_01,$ $\text{deforestation}]$ will represent deforestation\_-55.15\_-4.11\_2021\_08.png. To have a consistent naming convention for the date (\ie year\_month\_day), we add a nominal day label of “\_01” to all deforestation estimation test queries. For the test output, participants will submit in total 1,000 256$\times$256 binary masks, one 256$\times$256 binary mask for each input test query. Imagery related to the input test queries will be made available in our website and can be used to help generate the requested output.

\subsection{Deforestation Estimation Sub-Challenge}
\label{subsec:dec}
\noindent\textbf{Problem Definition.} 
Even after the forest fires are contained, their impact on the forest ecosystem can persist for years. Deforestation caused by fires and other factors remains a pressing issue in the Amazon. Automated deforestation estimation can help better understand the extent and rate of deforestation in the Amazon, identify areas that require intervention, and assess the effectiveness of conservation efforts. The goal of this sub-challenge is to perform a binary classification to predict whether a region is deforested or not. As solutions for the Fire Detection Sub-Challenge can be naturally extended,  we encourage participants in the Fire Detection Sub-Challenge to submit to the Deforestation Estimation Sub-Challenge. 

\noindent\textbf{Data.} Participants will use the multimodal remote sensing dataset described in Section~\ref{subsec:multimodal_data} to predict binary deforestation label maps.

\noindent\textbf{Metrics.} Performance is measured on based on pixel accuracy, F1 score, and Intersection over Union (IoU). 

\noindent\textbf{Submission Format.} Similarly, we provide separate guidelines for NetCDF and TIFF file users. \\
\noindent \textbf{(NetCDF File Users):} We will provide a NetCDF file that includes an array of 1,000 target test samples. Each sample will contain longitude, latitude, collection date, and a data source (‘Deforestation’) related to a test target. Functionality for retrieving images related to these test targets is provided in our code repository. \\ 
\noindent \textbf{(TIFF File Users):} We will provide a csv file that includes 1,000 test queries as a list of lists \ie [[${\text{lon}_0, \text{lat}_0, \text{date}_0, \text{modality}_0}],\text{…}, [\text{lon}_{999}, \text{lat}_{999}, \text{date}_{999}$, $\text{modality}_{999}]]$. Each test query is in the format [lon, lat, date, modality]. For example, $[-55.15, -4.11, 2021\_08\_01,$ $\text{deforestation}]$ will represent deforestation\_-55.15\_-4.11\_2021\_08\_01.tiff. To have a consistent naming convention for the date (\ie year\_month\_day), we add a nominal day label of “\_01” to all deforestation estimation test queries. Functionality for retrieving TIFF images related to these queries is provided in our code repository. 

Participants will submit in total 1,000 256$\times$256 binary masks, one 256$\times$256 binary mask for each input test query. 1's in the binary mask will represent deforested areas and 0's will represent forest/other areas. The output files should have dtype of uint8 and should use the naming convention Deforestation\_Lon\_Lat\_Date.tiff. For example, the expected output file name for test query [-55.15, -4.11, 2021\_08\_01, deforestation] is deforestation\_-55.15\_-4.11\_2021\_08\_01.tiff. Imagery related to the input test queries will be made available for the purpose of generating the requested output.

\subsection{Environmental Trend Prediction Sub-Challenge}
\label{subsec:mcc}

\noindent\textbf{Problem Definition.} Historical images obtained through remote sensing hold a vast amount of valuable information that can aid in understanding and predicting changes occurring on Earth. However, when dealing with humid forest areas such as the Amazon rainforest, acquiring clear images without cloud cover presents difficulties. In cases where traditional optical sensors are obstructed, SAR images offer an advantageous alternative to overcome the limitations imposed by clouds and provide enhanced imaging capabilities. The goal of the Environmental Trend Prediction Sub-Challenge is to forecast the visual appearance of the Amazon rainforest, regardless of weather and lighting conditions, by utilizing a combination of historical remote sensing data, including both multispectral and synthetic aperture radar images. To address this challenge, participants are encouraged to explore methods derived from representation learning, generative modeling, or any other viable approaches.

\noindent\textbf{Data.} Training set from the multimodal remote sensing dataset described in Section~\ref{subsec:multimodal_data} is provided. 

\noindent\textbf{Metrics.} To rank all submissions to this sub-challenge, imaging results entered by the participants will be evaluated using the following four metrics: Peak Signal-to-Noise Ratio (PNSR), Structural Similarity Index Measure (SSIM) \cite{ssim}, Learned Perceptual Image Patch Similarity (LPIPS) \cite{lpips}, and Fr\'{e}chet Inception Distance (FID) \cite{fid}. 

% To rank all submissions to the Matrix Completion Sub-Challenge, imaging results entered by the participants will be evaluated using the following four metrics: 1) Peak signal-to-noise ratio (PNSR) that computes in decibel (dB) the peak signal to noise ratio between two images; it is a quality measurement between an original image and a reconstructed version. The higher the PSNR, the better is the quality of the reconstructed image.  2) Structural similarity index (SSIM) \cite{ssim} quantifies the perceptual degradation of an image caused by processing such as data compression or losses that occurred in data transmission. 3) Learned perceptual image patch similarity (LPIPS) \cite{lpips} evaluates the distance between image patches, with a high value indicates dissimilarity and a low value means resemblance, and 4) Frechet inception distance (FID) \cite{fid} that assesses the quality of images created by a generative model.

\noindent\textbf{Submission Format.} We provide guildelines for users working with NetCDF and TIFF files separately. \\
\noindent \textbf{(NetCDF File Users):} We will provide a NetCDF file that includes an array of 2,000 target test samples. Each sample will contain longitude, latitude, collection date, data source, and band related to a test target. Functionality for retrieving images related to these test targets is provided in our code repository. \\ 
\noindent \textbf{(TIFF File Users):} We will provide a csv file that includes 2,000 test queries as a list of lists \ie [[${\text{lon}_0, \text{lat}_0, \text{date}_0, \text{modality}_0}],\text{…}, [\text{lon}_{1999}, \text{lat}_{1999}, \text{date}_{1999}$, $\text{modality}_{1999}]]$. Each test query is in the format [lon, lat, date, modality]. For example, $[-55.15, -4.11,  2021\_12\_04,$ $\text{Landsat8\_SR\_B2}]$ will represent Landsat8\_SR\_B2\_-55.15\_-4.11\_2021\_12\_04.tiff. Functionality for retrieving TIFF images related to these queries is provided in our code repository. 

Participants will submit in total 2,000 256$\times$256 images, one 256$\times$256 image for each input test query. The output files should have the following dtypes: float32 for Sentinel-1 bands and uint16 for Sentinel-2, Landsat 5 and Landsat 8 bands. The output files should use the naming convention Modality\_Lon\_Lat\_Date.tiff. For example, the expected output file name for test query [-55.15, -4.11, 2021\_12\_04, Landsat8\_SR\_B2] should be Landsat8\_SR\_B2\_-55.15\_-4.11\_2021\_12\_04.tiff. Imagery related to the input test queries will be made available and can be used to help generate the requested output.

\subsection{SAR-to-EO Image Translation Sub-Challenge}
\label{subsec:i2ic}

\noindent\textbf{Problem Definition.} The Amazon rainforest experiences high levels of cloud cover and frequent rainfall throughout the year. This leads to significant limitations for passive sensors which rely on sunlight to capture images. Clouds can obstruct the view of passive sensors, resulting in incomplete or distorted image data and gaps in the time series. Additionally, the Amazon rainforest is prone to wildfires, which can generate smoke and haze. These atmospheric conditions further reduce visibility and hinder the acquisition of clear and high-quality images. SAR is an active sensor that emits its own microwave signals and measures their reflections. SAR can penetrate through clouds and capture images even in adverse weather and lighting conditions, providing consistent data collection capabilities. However, SAR images often lack the intuitive visual interpretation offered by optical images, which possess rich color and fine-texture details. The goal of this sub-challenge is to model a distribution of possible electro-optical (EO) image outputs conditioned on a SAR input image. Here, EO image is a 3-channel RGB image from Sentinel-2. In Sentinel-2, RGB bands are represented as B4, B3, and B2, respectively. SAR image is a 2-channel Sentinel-1 image consisting of VV and VH bands.

% Obtaining a continuous time series of view of the Amazon rainforest is hindered by weather, clouds, smoke, and other inherent limitations of passive sensors (e.g. optical sensors) that rely on sunlight. Such limitations produce a major information gap in the Amazon rainforest that gets rain throughout the entire year. Synthetic aperture radar (SAR) is an active sensor that can collect images with relative invariance to weather and lighting conditions. However, visual interpretation of SAR images is not intuitive due to the large dynamic range, low spatial correlation, and radar-specific geometry distortion. To enhance SAR interpretability, this sub-challenge is aimed at modeling a distribution of possible electro-optical (EO) image outputs conditioned on a SAR input image. Here, EO image is a 3-channel RGB image from Sentinel-2. In Sentinel-2, RGB bands are represented as B4, B3, and B2, respectively. SAR image is a 2-channel Sentinel-1 image consisting of VV and VH bands. In this Image-to-Image Translation Sub-Challenge, participants need to model a distribution of potential results in a conditional generative modeling setting.

\noindent\textbf{Data.} Participants will use the multimodal remote sensing dataset described in Section~\ref{subsec:multimodal_data}. In our code repository, we provide a tool that can be used to retrieve aligned SAR and EO training images. The aligned dataset can be considered as $[\mathbf{x} ,[\mathbf{y}_1,\mathbf{y}_2,...,\mathbf{y}_N]]$ where a SAR image $\mathbf{x}$ is paired to a set of ground truth EO images $[\mathbf{y}_1,\mathbf{y}_2,...,\mathbf{y}_N]$. For each SAR image, all EO images of the same geographic region and which were collected within 7 days of the SAR image timestamp will be considered. On average 3 EO images are paired with each SAR image ($N \approx 3$).

% For this sub-challenge, we provide JSON files specifying which Sentinel-2 EO images (B4, B3, and B2) correspond to which SAR images. We provide two JSON files: one for Sentinel-1 VV band\footnote{\url{https://rainforestchallenge.blob.core.windows.net/dataset/sentinel_vv_image_alignment_train.json}} and another for Sentinel-1VH band \footnote{\url{https://rainforestchallenge.blob.core.windows.net/dataset/sentinel_vh_image_alignment_train.json}}. The mappings in both files are identical. The aligned dataset will have the following format: $[\mathbf{x} ,[\mathbf{y}_1,\mathbf{y}_2,...,\mathbf{y}_N]]$ where a SAR image $\mathbf{x}$ is paired to a set of ground truth EO images $[\mathbf{y}_1,\mathbf{y}_2,...,\mathbf{y}_N]$. For each SAR image, all EO images of the same geographic region and which were collected within 7 days of the SAR image timestamp will be identified.  On average 3 EO images are paired with each SAR image ($N \approx 3$). 
% % \phillip{Should we describe here how the $N$ ground truth images are defined? People may wonder why there isn't just a single ground truth.}

\noindent\textbf{Metrics.} Performance is evaluated based on following metric:
\begin{equation}
\sum_j \min_i \Arrowvert f(\mathbf{x})_i - \mathbf{y}_j \Arrowvert
\end{equation}
 where $f(\mathbf{x})_i$ is a set of possible EO images translated from  a generative model $f(\cdot)$ conditioned on an input SAR image $\mathbf{x}$, and  $\mathbf{y}_j$ is an EO image from the corresponding ground truth set. This metric also evaluates diversity of generated output images staying faithful to the diversity of the ground-truth EO data. 

\noindent\textbf{Submission Format.} Participants will submit in total 15,000 256$\times$256$\times$3 EO images. Three translated 256$\times$256$\times$3 EO images for each input Sentinel-1 SAR image. \\
\noindent \textbf{(NetCDF File Users):} We will provide a NetCDF file that includes an array of 5,000 test samples. Each sample will contain longitude, latitude, collection date, data source (‘Sentinel-2’), and bands (B4, B3, B2) related to a test target. Each target sample will have a collection date exactly corresponding to the corresponding Sentinel-1 SAR image that serves as an input. Our example code provides functionality for retrieving images related to these test targets.\\ 
\noindent \textbf{(TIFF File Users):} 10,000 single-pol SAR images will be provided. The SAR images will have shape 256$\times$256$\times$1 with 5,000 having VV polarization and the other 5,000 the corresponding VH polarization. 

This is a multimodal image-to-image translation problem where participants will generate three possible EO images given an input SAR image. The output files should have dtype of uint16 and should use the naming convention SAR2EO\_Modality\_Lon\_Lat\_Date\_Sample\#.tiff. For example, For example, the expected output file names for input SAR images, Sentinel1\_VH\_-54.80\_-4.01\_2019\_03\_18.tiff and Sentinel1\_VV\_-54.80\_-4.01\_2019\_03\_18.tiff should be \\SAR2EO\_Sentinel2\_EO\_-54.80\_-4.01\_2019\_03\_18\_1.tiff, SAR2EO\_Sentinel2\_EO\_-54.80\_-4.01\_2019\_03\_18\_2.tiff, and SAR2EO\_Sentinel2\_EO\_-54.80\_-4.01\_2019\_03\_18\_3\\.tiff. Additional SAR imagery  related to these tests will be made available for the purpose of generating the requested output.

% For testing, we will provide 5,000 Sentinel-1 SAR images, where each SAR image is 256$\times$256$\times$2 and the 2 channels correspond to the Sentinel-1 VV and VH bands, respectively. For the test output, participants will submit in total 15,000 256$\times$256$\times$3 EO images, three translated 256$\times$256$\times$3 EO images for each input Sentinel-1 SAR image. The EO image channels correspond to the Sentinel-2 RGB channels (B4, B3, B2). This is a multimodal image-to-image translation problem where participants will generate three possible EO images given an input SAR image. 

\section{Conclusion}
In this paper, we introduce MultiEarth 2023 – the second Multimodal Learning for Earth and Environment Challenge. It comprises four sub-challenges: fire detection, deforestation estimation, environmental trend prediction, and SAR-to-EO image translation. This paper outlines the challenge problems, data sources, evaluation metrics, and submission guidelines. For the challenge, we have curated a multimodal dataset that features a continuous time series of Sentinel-1, Sentinel-2, Landsat 5 and Landsat 8, with aligned fire and deforestation labels. In addition, we provide a comprehensive documentation of the dataset and an API along with a tutorial that helps researchers the data processing and filtering.

\section{Acknowledgments}
% The authors would like to thank Dr. M.K. Newey for helping the team labeling the August 2021 time slice data, and Scale AI, for its efforts to label, in a short period of time, the remaining ten time slices considered in this MultiEarth2022.
\scriptsize
Research was sponsored by the United States Air Force Research Laboratory and the United States Air Force Artificial Intelligence Accelerator and was accomplished under Cooperative Agreement Number FA8750-19-2-1000. The views and conclusions contained in this document are those of the authors and should not be interpreted as representing the official policies, either expressed or implied, of the United States Air Force or the U.S. Government. The U.S. Government is authorized to reproduce and distribute reprints for Government purposes notwithstanding any copyright notation herein.

%%%%%%%%% REFERENCES
{\small
\bibliographystyle{ieee_fullname}
\bibliography{paper}
}

\end{document}